\documentclass{llncs}
\usepackage{makeidx}  
\usepackage{graphicx}
\usepackage{float}
\usepackage{caption}
\usepackage{subcaption}
\begin{document}
%
\mainmatter              
\title{Shopper Analytics: a customer activity recognition system using a distributed RGB-D camera network}
\titlerunning{Shopper Analytics}  
%
\author{Daniele Liciotti  \and Marco Contigiani \and Emanuele Frontoni \and Adriano Mancini \and Primo Zingaretti\inst{1} \and Valerio Placidi\inst{2}}
\authorrunning{Emanuele Frontoni et al.} 
%
\tocauthor{Daniele Liciotti, Marco Contigiani, Emanuele Frontoni, Adriano Mancini, Primo Zingaretti and Valerio Placidi}
\institute{Dipartimento di Ingegneria dell'Informazione, Universit\`{a} Politecnica delle Marche, Via Brecce Bianche, 60131 Ancona, Italy,\\
\email{\{d.liciotti, m.contigiani,e.frontoni, a.mancini, p.zingaretti\}@univpm.it}\\ 
\and Grottini Lab srl,\\
 Via S.Maria in Potenza, 62017, Porto Recanati, Italy,\\
\email{valerio.placidi@grottinilab.com}}
\maketitle              
\begin{abstract}
The aim of this paper is to present an integrated system consisted of a RGB-D camera and a software able to monitor shoppers in intelligent retail environments. We propose an innovative low cost smart system that can understand the shoppers' behavior and, in particular, their interactions with the products in the shelves, with the aim to develop an automatic RGB-D technique for video analysis. The system of cameras detects the presence of people and univocally identifies them. Through the depth frames, the system detects the interactions of the shoppers with the products on the shelf and determines if a product is picked up or if the product is taken and then put back and finally, if there is not contact with the products. The system is low cost and easy to install, and experimental results demonstrated that its performances are satisfactory also in real environments.\\
\keywords{marketing retail, consumer behavior, integrated architecture, RGB-D camera, video analysis}
\end{abstract}
\section{Introduction}
In the last years, the analysis of the human behavior has been of high interest to researchers because its important and different applications, such as: video surveillance \cite{ko,cristani}, ambient assisted living \cite{frontoni1}, analysis of consumer's behavior \cite{frontoni2}, group interactions and many others. In particular, in the field of intelligent retail environments, numerous studies to investigate how shoppers behave inside a store and how businesses can change strategies to improve sales are emerging. In order to analyse the buyer activity and to solve general aspects of these problems, techniques of artificial intelligence are used and, in particular, vision and image processing. In recent years, the visual analysis of dynamic scenes is one of the most important research activities in computer vision and image understanding~\cite{ascani,desai,ferrari}. When the visual analysis concerns moving scenes, the general method includes following steps: modelling of environments, motion detection, human identification, classification of moving objects, tracking, behavior understanding and data fusion from multiple cameras \cite{desai,brox,farhadi}.
In this manuscript, we focus the attention on the study of the consumer behavior in a real retail store, in order to recognize human actions \cite{burdev,csurka,felzenszwalb,gupta}, such as ``interacting with the shelf'', ``picking or releasing a product'', ``moving in a group'', and ``knowing most visited areas in the store''. Consumers are main actors in the project because the goal is to increase their satisfaction and, therefore, enhance their purchases.
Currently, the identification of the shoppers' behavior implements systems of human observation or video recording with traditional cameras. Some tools, such as virtual stores or eye tracking provide incomplete and unrepresentative data because they are based on a small sample of buyers.
As a result, by univocally identifying shoppers and automatically analysing their interactions with the products on the shelves and their activities in different zones, our design considerably increases the value of the current marketing research methodologies. Moreover, the main innovation concerns the original use of tracking system, and the other interesting point concerns the real experimental platform described in the results section combined with a vision based statistical approach.
Therefore, the project aims to propose an intelligent low-cost embedded system able to univocally identify customers, to analyse behaviors and interactions of shoppers and to provide a large amount of data on which to perform statistics. The automatic extraction of features that univocally recognize each subject in the scene and their movements, provides an important tool to identify important operations concerning marketing strategies.
The application implements techniques of image processing such as: background subtraction, low-level segmentation, tracking and finding contours, in order to map a single shopper and/or a group of people within the store that interact with the products on the shelves, defining an ID unique to each visitor filmed by the camera, and classifying these interactions.
Paper is organized as following described: Section~\ref{retail} introduces the main aspects of marketing retail and consumer behavior. Section~\ref{arch} in detail describes the architecture of the system. The experimental setup is described in Section~\ref{setup} and the results are showed in Section~\ref{results}. Last Section~\ref{conc} described conclusions and future works.

\section{Marketing Retail and Consumer Behavior}
\label{retail}
The need to associate marketing retail and consumer behavior is born from the necessity to develop theories, strategies and management models compatible with customer behavior. The concept of shop is changed during years becoming not only the place where customers go to buy a specific product, but also the place where the customers go to spend part of their time. Therefore, it is very important to study the consumer behavior so as to investigate the elements of the decision-making process of purchase that determines a particular choice of consumers and how the marketing strategies can influence the customer. 
Empirical researches on consumer behavior are primarily based on the cognitive approach, which allows to predict and define possible actions that lead to the conclusion and to suggest implications for communication strategies and marketing. The basic principle of this approach is that individual actions are the result of information processing. The person collects the information, interprets, processes and uses them to take action. Cognitive approaches cannot completely explain the complexity of consumer behavior, which lives in a changing social and cultural context. According to this approach, the choice of purchasing comes from the ability of the products to generate specific sensations, images and emotions.
According Perreau~\cite{perreau}, five are the steps of consumer buying decision process:
\begin{enumerate}
\item \textit{Perception of the problem}: the shopper recognizes a gap between the current situation and desirable situation, therefore perceives a need. The need can be described as a genuine request that comes from the inside and the satisfaction of which is necessary for the survival or to maintain a good level of psychophysical balance.
\item \textit{Research of information}: in order to identify the satisfying solution for the perceived need, the consumer searches for knowledge in the memory, or if the information possessed by the individual is not sufficient, will seek additional data from external sources.
\item \textit{Evaluation of options}: consists of selecting one of the available alternatives based on the criteria defined in the previous step.
\item \textit{Buying decision}: after having identified the place and time.
\item \textit{Post purchase behavior}: is the adequacy of the product purchased and thus the level of consumer satisfaction.
\end{enumerate}
Therefore, the marketing retail discipline defines the set of marketing strategies to point of sale oriented so as to attract the customer and to increase the activities of businesses. To achieve its objectives, the retail marketing uses many techniques through several stages of planning by developing a marketing model for the shop-customer using the most important techniques, in the following described:
\begin{itemize}
\item Visual merchandising is the activity of developing floor plans in order to maximize sales. The purpose is to attract, engage and motivate the shopper towards making a purchase. As means of visual merchandising is often widely used a planogram \cite{mankodiya}.
\item Pricing is the activity of establishing the best price that is competitive for shoppers and at the same time with a good profit margin for the store.
\item Sensory marketing, to make the shopping experience more pleasant and exciting for the client.
\item Loyalty tools, to encourage the consumer to return to the store and to make new purchases.
\item Non-conventional marketing concerns original ideas to push the customer to come into the store and trigger a word of mouth process.
\end{itemize}
The best way to know the behavior of the customer is to create an automatic system that, on the base of acquired knowledge, can predict the purchase of many products and also choices. Therefore, the first goal of this work is to assign an ID unique to each person detected by a vertical mounted RGB-D camera, to track their activity within the store and then to detect their interactions with the shelf. The next step is to analyse and to classify the interactions: indicating if the product has been picked up and purchased or if the product has been put back after picked up. So, the proposed system will identify the activity of the consumer in front of the shelf.

\section{Overview of the System Architecture}
\label{arch}
In order to satisfy both functional and non-functional requirements of the system, a Single Board Computer (for example, Raspberry Pi) has been used, since it is sufficiently small and suited to manage all functions. Functional requirements are: counting and classification of people, their interaction with the shelf, sending data to web server and data analysis; while non-functional requirements are: place of installation and connection modes. As RGB-D sensor, Asus Xtion Pro live has been chosen due to its smaller dimensions than Microsoft Kinect, and the power supply is provided only by USB port. It does not need an additional power.

\begin{figure}[t]
\centering
\includegraphics[width=9cm]{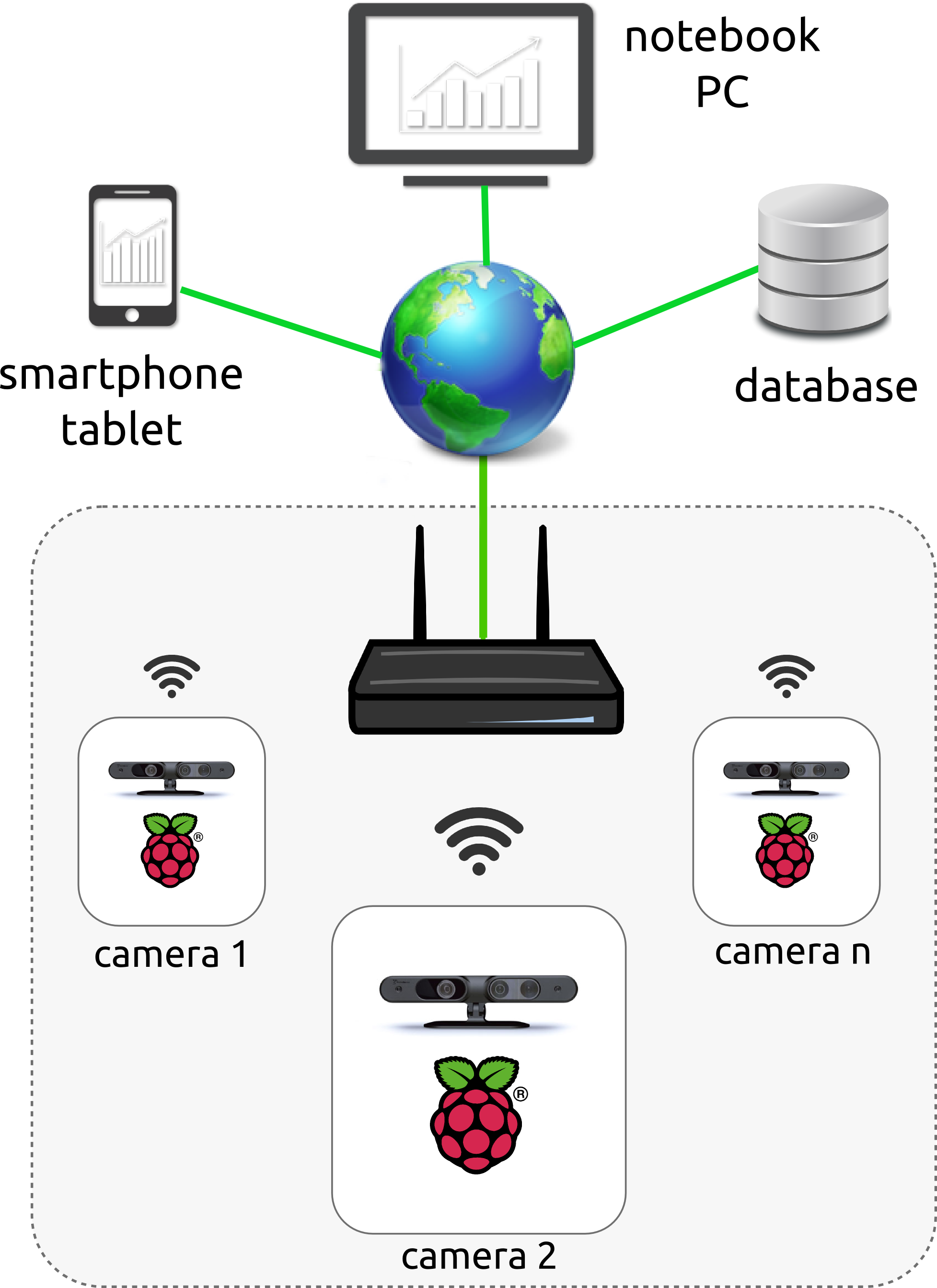}
\caption{General scheme of connection between the components.}\label{general-scheme}
\end{figure}

Figure \ref{general-scheme} shows the general scheme of the implemented system and the interactions between the components. The system consists of six devices, listed below:

\begin{enumerate}
\item Single Board Computer: is a complete computer built on a single circuit board, with microprocessor(s), memory, input/output (I/O) and other features required of a functional computer. Single-board computers were made as demonstration or development systems, for educational systems, or for use as embedded computer controllers.
\item Asus Xtion Pro live: is composed by an infrared sensor, a RGB sensor and 2 microphones. It is able to provide in output a RGB representation of the scene and also allows to reconstruct a depth map of the same. In the depth map the value of each pixel codifies the distance of each element from 3D scene.
\item Wireless Adaptator. 
\item SD/MicroSD Memory Card 8GB Speed 10.
\item Hub USB 2.0: has the task of ensuring the supply of the RGB-D sensor. 
\item Router 3G/4G Wireless.
\end{enumerate}

The Single Board Computer uses a SD memory card where Debian operating system is installed allowing an easy configuration of RGB-D sensor of Asus Xtion Pro Live compiling following modules: OpenNI Library\footnote{https://github.com/OpenNI/} and PrimeSense Sensor Driver\footnote{https://github.com/PrimeSense/Sensor}. 
\begin{figure}[t]
\centering
\includegraphics[width=10cm]{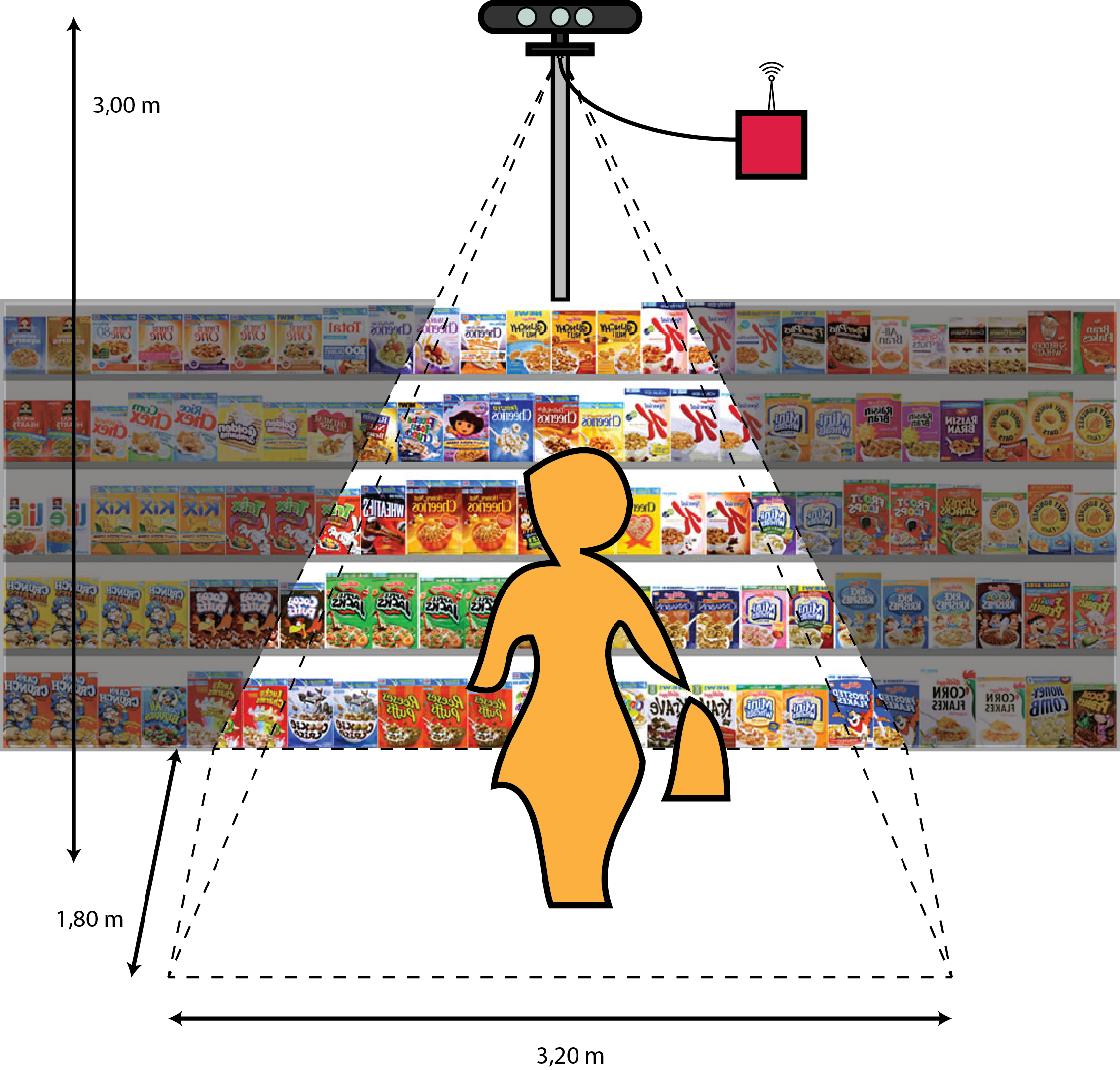}
\caption{Environment of system installation.}\label{view-system}
\end{figure}

\begin{figure}[t]
\centering
\includegraphics[width=7cm]{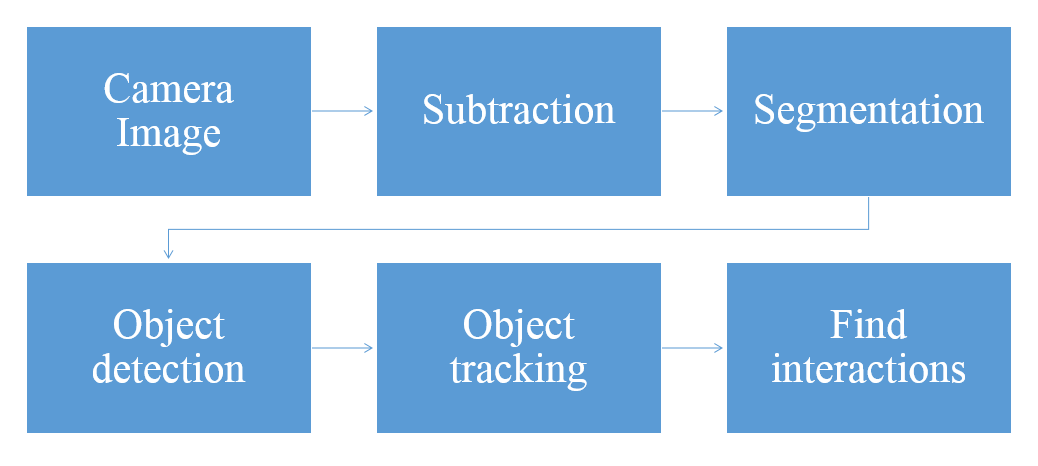}
\caption{Logical steps of the implemented algorithm.}\label{steps}
\end{figure}

The RGB-D sensor is installed (Fig. \ref{view-system}) in a top view configuration at three meters of height from the floor. It visualizes a maximum area (shopper tracking area) of 1.8m x 3.2m, but the shelf area (shelf tracking area), that has a height of two meters, results smaller than this. The system implements the algorithm that calculates the interactions map between the people in the store and the shelf, sending successively data to a database. Trough a PC, it is possible to connect a smartphone to the database and to visualize the state of system and other interesting information.
Figure \ref{steps} represents the block diagram that identifies the main steps of the algorithm. The input is the image detected by the camera and the output is the typology of interaction between the user and the products on the shelf.

\begin{figure}[h]
\centering
\includegraphics[width=6cm]{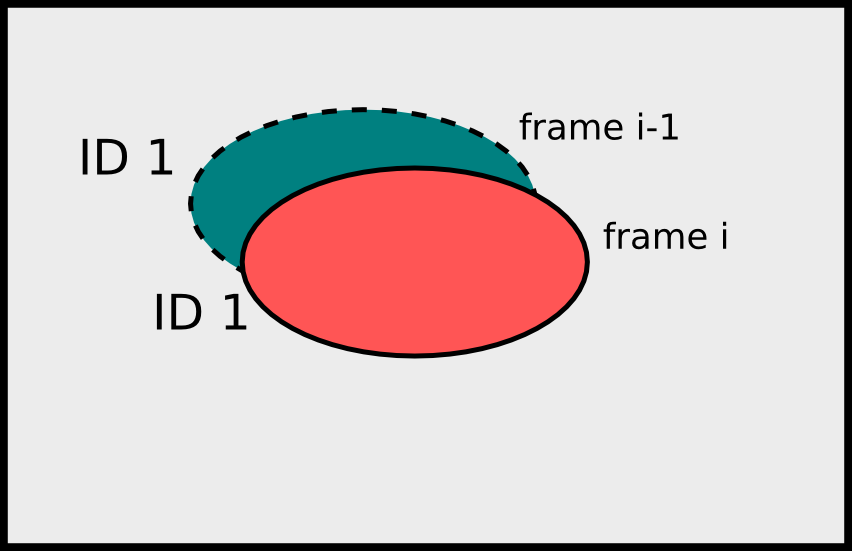}
\caption{Object tracking implementation.}\label{frame}
\end{figure}

In the first step, the system acquired the streaming video from the RGB-D sensor. After this, the background subtraction method is implemented, that is one of the most commonly used algorithms for detection of moving objects within a sequence of images. 
This approach is reliable since each pixel also maintains the depth information, that is not available with a RGB image and so it allows to detect the distance of each blob. Moreover, in order to avoid false detection of objects (false positives), the background image is dynamically updated. After the background subtraction, a threshold value is defined that allows to discriminate positive signals that indicate moving objects, by false positives due to background noise, this method is called segmentation. Another important step consists of the object detection where, for each significant blob, the boundary and the maxima points are found, corresponding to the head of the person. If these points are surrounded by a region of the lowest points comparable to jump head-shoulder of a human then is a valid blob \cite{migniot}. The next phase is the object tracking that recognizes the pathways of different blobs along the frames. In other words, in this phase, each blob is recognized and tracked within the streaming video. For each person, the height is determined verifying that this is in the neighborhood of the height of the person in the previous frame. This method is easy but very effective since it is based on the depth image; moreover it is not subject to rapid changes in the forms, allowing a good and reliable tracking. Figure \ref{frame} shows how the people are tracked between two successive frames (frame \textit{i-1} and frame \textit{i}). In both frames, the same identifier ($ID_{1}$) detects the same blob, tracked between frames, so each identifier univocally identifies a person. In this phase of the work, users are not tracked across the sensors, but we retain that this approach must be investigated in future, so that to each visitors maintains a ID unique during the entire visit to the store.

The last step of the algorithm provides the find interactions procedure. When a person has a contact with the shelf, the associate blob is inside the shelf zone. Then, it is possible to detect the exact point of contact by means the definition of common 3-dimensional (XYZ) system coordinates.

The shelf zone, that is defined by user in a configuration file, is formed by three parameters (x shelf dist sx, x shelf dist dx and y shelf dist) as also showed in the following figure \ref{shelfzone}. 

\begin{figure}[h]
\centering
\includegraphics[width=12cm]{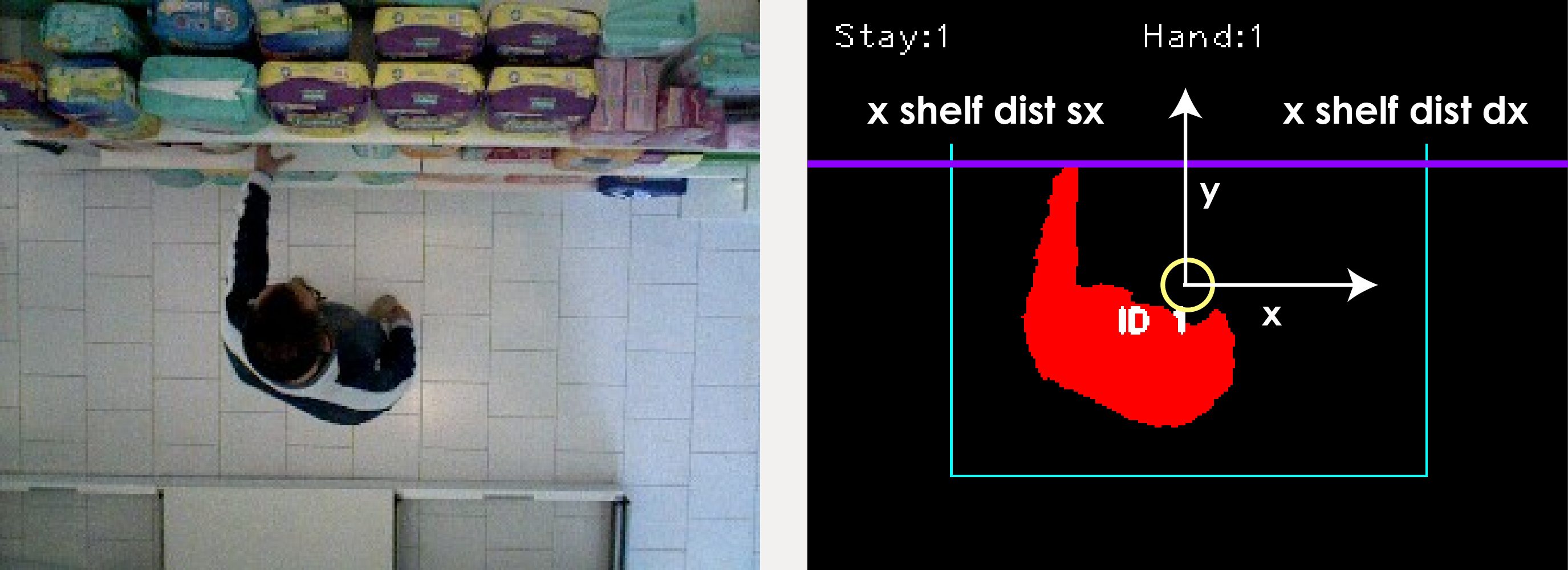}
\caption{Setting parameters of the shelf zone.}\label{shelfzone}
\end{figure}

When the people interact with the shelf can be presented three different situations, classified as follows:

\begin{enumerate}
\item \textit{Positive}: when the product is picked up from the shelf.
\item \textit{Negative}: when the product is taken and then repositioned on the shelf.
\item \textit{Neutral}: if the hand exceeds the threshold without taking anything.
\end{enumerate}

The template matching method has been used to identify and to classify the interactions between the people and the shelf. So, when there is the first contact, the position of the hand in the RGB image is saved, and the same operation occurs when the interaction ends, in order to compare the first image and the final image. If there is a significant correspondence, the interaction is neutral, since there is not an important difference between the first and final image. Otherwise, the interaction can be positive or negative. To identify the type of interaction, the area of the blobs, that is present in the contours, between the two images has been considered.

\subsection{Interactions Map}
\begin{figure}[t]
\centering
\includegraphics[width=7cm]{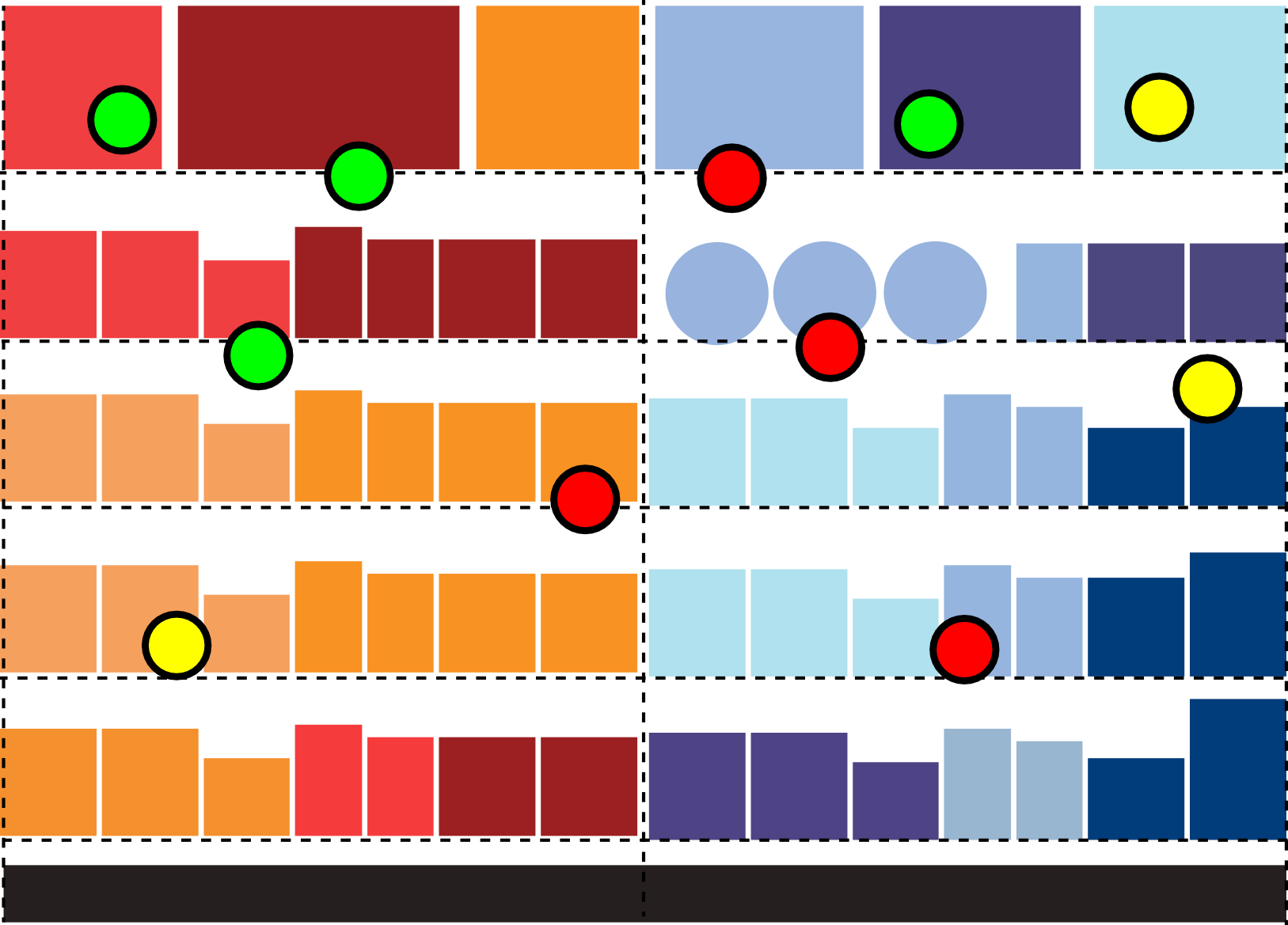}
\caption{Interactions Map produced by the software.}\label{heatmap}
\end{figure}

This architecture implements a function that displays an interactions map on the screen, during the execution of the program. This is very important because in real time is possible to view the information in which area of the shelf there have been contacts. Figure \ref{heatmap} shows an example of the interactions map. This function draws a colored ball corresponding to the point of contact, on a planogram previously loaded by the user. The color of the ball depends on the type of interaction (green=positive, red=negative and yellow=neutral). This function is used during the debugging phase, in fact all data are saved in a database for later analysis.

\section{Experimental setup}
\label{setup}
The development of the procedure to verify the performances of the system has been realized thanks to the collaboration of Grottini Lab that provided the material and, moreover, allowed to test the system in their laboratory and successively in a real store, partner of Grottini Lab.
The collaboration with the partner has been very useful to decide the arrangement of the system, according to functional strategic locations for sales and for the input monitoring.
All the system has been installed on a panel in the suspended ceiling of the store. Each system gives in output a significant amount of data that are stored in a database, so that they can be successively analyzed to extract indicators.
The final test in the real store has been realized installing four RGB-D cameras for a time period of three months, in order to obtain significant and real data. The cameras monitored the entrance (camera 1), the bleach zone (camera 2), the perfumes zone (camera 3) and the shampoo zone (camera 4). The choice to put a camera near the entrance allowed to exactly count the number of people who entered the store. The indicators that are useful to evaluate the shopper behavior and that can help the store staff to understand their preferences and finally, to increase the sales, are:

\begin{itemize}
\item Total number of visitors;
\item Total number of shoppers;
\item Number of visitors in a particular zone;
\item Number of visitors interacting with the shelf;
\item Number of interactions for each person;
\item Number of visitors becoming shoppers (sales conversion);
\item Average visit time.
\end{itemize}
Some indicators that consider the interactions can be:
\begin{itemize}
\item Number of products picked up;
\item Number of products relocated on the shelf;
\item Number of products touched;
\item Duration of interactions;
\item Average interaction time;
\item Number of interactions for product and for category.
\end{itemize}

\section{Results}
This section presents some experimental results aimed at highlighting the performances of the system in a real environment. The system extracts a high number of parameters, we retain that the most significance to show how it behaves in a real situation are presented in the following graphs.

\label{results}
\begin{figure}	
	\centering
	\begin{subfigure}[t]{5cm}
		\centering
		\includegraphics[width=5cm]{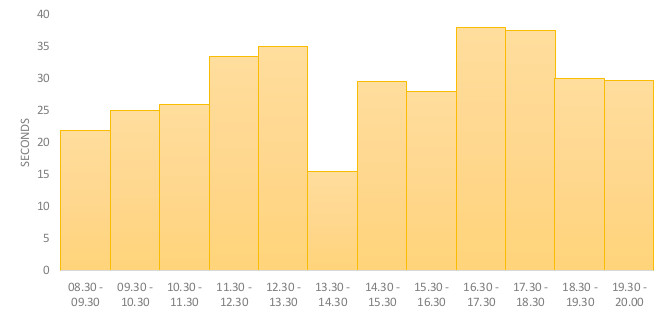}
		\caption{Average visit time [sec] in the bleach zone for each hour.}\label{fig:1a}		
	\end{subfigure}
	\quad
	\begin{subfigure}[t]{5cm}
		\centering
		\includegraphics[width=5cm]{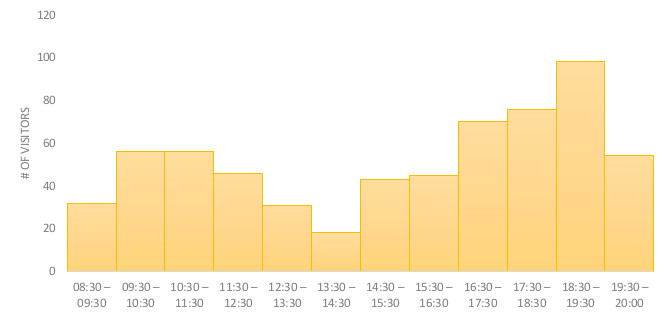}
		\caption{Number of visitors in the bleach zone for each hour.}\label{fig:2b}
	\end{subfigure}
		\quad
	\begin{subfigure}[t]{5cm}
		\centering
		\includegraphics[width=5cm]{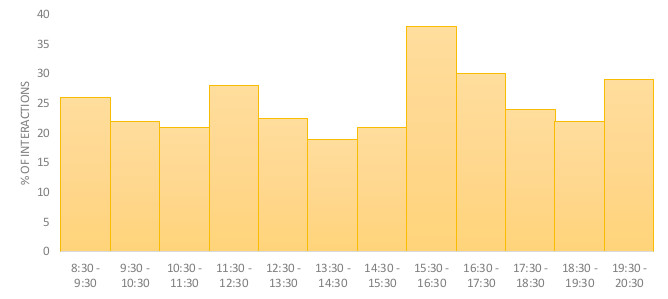}
		\caption{Percentage of interactions in the bleach zone for each hour.}\label{fig:3c}
	\end{subfigure}
	\caption{Some graphics used to evaluate the shopper behavior.}\label{fig:1}
\end{figure}

In the three figures we showed the average trend of some important parameters referred to a day of a week. Figure \ref{fig:1a} shows the trend of the average time in seconds of visitors in the bleach zone. Observing the graph, the number of visitors is low during the early hours of day (from 8:30 to 9:30) and also during the central hours of day (from 12:30 to 14:30) probably due to less time for making purchases because closer to working time. The other graph that highlights an interesting data is showed in Fig. \ref{fig:2b}, which it indicates the trend of the number of visitors always in the bleach zone. From 12:30 to 13:30, in that area, the visitors are 20 on average. This data implies a review of sales strategies, through alternatives marketing solutions as, for example, a more pronounced sign or a temporary closure during the lunchtime.
The last figure \ref{fig:3c} indicates the percentage of visitors that have had at least one interaction with the shelf where the bleach is located. Visitors that have picked up a product or have had a contact with the shelf have been 27\% on average. In order to evaluate the performance of the integrated system that we realized, we compared the number of interactions detected by the system and the real interactions physically determined, by obtaining a reliability factor near to 96\%. The system quite correctly recognize if a visitor crosses along the zone without any stop in front of the shelf, or if he interacts with it. Moreover, the interactions map gives additional visual information on the typology of action that the visitor performed becoming a potential buyer.

\section{Conclusion and Future Works}
\label{conc}
In this work, the goal is to propose an automatic and intelligent system able to analyse and to classify the behavior of customers within a retail store. This system received many interest of retailers because, since from the extracted data, they can derive useful information about the behavior of the customer in front of a shelf. With this technology, it is possible to measure which area of the shelf attracts the attention of the customer, in which shelf to place the products in launch phase, how long the customer remains opposite the shelf and which areas are most visited. From the indicators extracted from the system the retailers can employ a number of marketing strategies in order to attract the customers attention. In the experimental phase, the test of the system in a real environment has provided very interesting results. In particular the architecture is stable, easy to install and especially convenient due to the low cost components. The data provided by the system, that are stored in a database, are very reliable and responsive to the real situation. In fact, in the analysing phase the data, when the interaction is positive, it is possible to correctly identify the product that the customer has picked up. The system ensures a rather high reliability, especially in an ideal condition in which the shelves are those considered to arm height. In the future, the accuracy of the system will be improved independently by the position of the product on the shelf and the consumer position. In addition, the optimization of the image processing algorithm is required, in order to implement an effective procedure of detection and tracking of the shoppers in different areas of the store.

%
%

\clearpage
\addtocmark[2]{Author Index} 
\renewcommand{\indexname}{Author Index}
\printindex
\clearpage
\addtocmark[2]{Subject Index} 
\markboth{Subject Index}{Subject Index}
\renewcommand{\indexname}{Subject Index}

\begin{thebibliography}{100}
	\bibitem{ko} T.~Ko, Co.~Raytheon and Va.~Arlington, ``A survey on behavior analysis in video surveillance for homeland security applications'', Applied Imagery Pattern Recognition Workshop, 2008. AIPR '08. 37th IEEE, pp. 1-8,  15-17 Oct. 2008.
	\bibitem{cristani} M.~Cristani, R.~Raghavendra, A.~Del Bue, and V.~Murino, ``Human behavior analysis in video surveillance: A Social Signal Processing perspective'', Neurocomputing vol.100 Special issue: Behaviours in video, pp. 86-97, 16 January 2013.
	\bibitem{frontoni1} E.~Frontoni, A.~Mancini and P.~Zingaretti, ``RGBD Sensors for human activity detection in AAL environments'', Living Italian Forum 2013 Longhi, S., Siciliano, P., Germani, M., Monteri\'u, A. (Eds.), 300 p. 50 illus. Available Formats: eBook ISBN 978-3-319-01118-9, Due: July 31, 2014.
	\bibitem{frontoni2} E.~Frontoni, P.~Raspa, A.~Mancini, P.~Zingaretti and V.~Placidi, ``Customers' activity recognition in intelligent retail environments'' Lecture Notes in Computer Science (including subseries Lecture Notes in Artificial Intelligence and Lecture Notes in Bioinformatics), 8158 LNCS, pp. 509-516, 2013.
	\bibitem{ascani} A.~Ascani, E.~Frontoni, A.~Mancini and P.~Zingaretti, ``Feature group matching for appearance-based localization'', IEEE/RSJ 2008 International Conference on Intelligent RObots and Systems Ð IROS 2008, Nice, 2008.
	\bibitem{ferrari} V.~Ferrari, M.~Marin-Jimene, and A.~ Zisserman, ``Pose search: retrieving people using their pose'', International Conference on Computer Vision and Pattern recognition IEEE/CVPR, pp. 1-8, 2009.
	\bibitem{desai} C.~Desai, D.~Ramanan, and C.~Fowlkes, ``Discriminative models for static human-object interactions'', Computer Vision and Pattern Recognition Workshops (IEEE/CVPRW), pp. 9-16, 2010.	
	\bibitem{brox} T.~Brox, L.~Bourdev, S.~Maji and J.~Malik, ``Object segmentation by alignment of poselet activations to image contours'', International Conference on Computer Vision and Pattern recognition IEEE/CVPR, pp. 2225-2232, 2011.
	\bibitem{farhadi} A.~Farhadi, I.~Endres, D.~Hoiem and D.~Forsyth, ``Describing objects by their attributes'', International Computer on Vision and Pattern Recognition, IEEE/CVPR, pp. 1778-1785, 2009.
	\bibitem{burdev} L.~Bourdev and J.~Malik, ``Poselets: Body part detectors trained using 3D human pose annotations'', International Conference on Computer Vision IEEE/ICCV, pp. 1365-1372, 2009.
	\bibitem{csurka} G.~Csurka, C.~Bray, C.~Dance and L.~Fan, ``Visual categorization with bags of keypoints''. Workshop on Statistical Learning in Computer Vision'', ECCV, pp. 1-22, 2004.
	\bibitem{felzenszwalb} P.~ Felzenszwalb, R.~ Girshick, D.~ McAllester and D.~ Ramanan, ``Object detection with discriminatively trained part based models'', IEEE Transactions on Pattern Analysis and machine Intelligence, vol. 32, no. 9, pp. 1627-1645, 2010.
	\bibitem{gupta} A.~Gupta, A.~Kembhavi, and L.~Davis, ``Observing human-object interactions: Using spatial and functional compatibility for recognition'', IEEE Transactions on Pattern Analysis and machine Intelligence vol. 31, no. 10, pp. 1775-1789, 2009.
	\bibitem{perreau} F.~Perreau, ``The forces that drive consumer behavior and how to learn from it to increase your sales'', 2013, theconsumerfactor.com.
	\bibitem{mankodiya} K.~ Mankodiya, R.~Gandhi, and P.~Narasimhan, ``Challenges and Opportunities for Embedded Computing in Retail Environments'', Lecture Notes of the Institute for Computer Sciences, Social Informatics and Telecommunications Engineering, vol. 102, pp 121-136, 2012.
	\bibitem{migniot} C.~Migniot and F.~Ababsa, ``3D Human Tracking from Depth Cue in a Buying Behavior Analysis Context'', Lecture Notes of the Institute for Computer Sciences, Computer Analysis of Images and Patterns, vol. 8047, pp. 482-489, 2013.
\end{thebibliography}
\end{document}